# Stochastic phonological grammars and acceptability

**John Coleman**
Phonetics Laboratory
University of Oxford
United Kingdom
john.coleman@phon.ox.ac.uk

**Janet Pierrehumbert**
Department of Linguistics,
Northwestern University, and
Département Signal, ENST, Paris
jbp@nwu.edu

## Abstract

In foundational works of generative phonology it is claimed that subjects can reliably discriminate between possible but non-occurring words and words that could not be English. In this paper we examine the use of a probabilistic phonological parser for words to model experimentally-obtained judgements of the acceptability of a set of nonsense words. We compared various methods of scoring the goodness of the parse as a predictor of acceptability. We found that the probability of the worst part is not the best score of acceptability, indicating that classical generative phonology and Optimality Theory miss an important fact, as these approaches do not recognise a mechanism by which the frequency of well-formed parts may ameliorate the unacceptability of low-frequency parts. We argue that probabilistic generative grammars are demonstrably a more psychologically realistic model of phonological competence than standard generative phonology or Optimality Theory.

## 1 Introduction

In standard models of phonology, the phonological representation of a word is understood to be a hierarchical structure in which the phonological material (features and/or phonemes) is organized into syllables, which are in turn organized into feet, prosodic words and intonation phrases. The existence of such structure is supported by a confluence of evidence from phonotactic constraints, patterns of allophony, and results of psycholinguistic experiments. In this paper, we present a probabilistic phonological parser for words, based on a context free grammar. Unlike classical probabilistic context-free grammars (Suppes 1972), it attaches probabilities to entire root-to-frontier paths instead of to individual rules. This approach makes it possible to exploit regularities in the horizontal, or time-wise, location of frequency effects. The grammar is applied to model phonological productivity as revealed in acceptability ratings of nonsense words. Specifically, we examine the issue of whether acceptability is related to expected frequency as computed over the whole word (with deviations in different locations having a cumulative effect), or whether the judgments of acceptability are dominated by the local extreme values. We find that an experimentally obtained measure of subjective phonotactic "badness" correlates with three probabilistic measures: word probability, log word probability, and frequency of the lowest frequency (i.e. "worst") constituent.

The hierarchical structures of phonology obviously lend themselves to being formalized using standard types of grammars. Formalization makes it possible to rigorously relate generation and parsing. It allows us to test particular linguistic theories of prosody by evaluating their predictions over large data sets. Previous work which has established these points includes Church (1983), Randolph (1989), and Coleman (1992).

Prosodic structure in some respects presents a simpler problem than syntactic structure, because the inventory of different node types is small and the grammar lacks recursion. In terms of weak generative capacity, the grammar can obviously be treated as finite state. The linguistically transparent prosodic grammar presented in this paper was developed for the purpose of modeling phonological productivity. The grammar is trained on an existing dictionary, and it is applied to model judgments of well-formedness obtained for a study of the psychological reality of phonotactic constraints (Coleman 1996). For the study, nonsense words were constructed which either respected or violated known phonotactic constraints, and subjects indicated by pressing one of two buttons whether or not the nonsense word could be a possible English word. The total number of "votes" against each word, from 6 subjects on 2 runs yields a scale of 0 (good) to 12 (bad). For example, the nonsense word /ˈsmlɒfɪt/

contains an extremely anomalous onset cluster, and it received 10 votes against. In contrast, the nonsense word /ˈtɹɛlm/ did not violate any known phonotactic constraints, and it received only 2 votes against.

We undertake to model productivity because it is a standard diagnostic for the psychological reality of abstractions. Modeling in detail the perceived well-formedness of neologisms provides us with an opportunity to assess how prosodic structure figures in the cognitive system. Although earlier work has established a connection between lexical statistics and acceptability, no general architecture for manipulating lexical statistics in a structure-sensitive fashion has yet been developed.

The connection between lexical statistics and acceptability is demonstrated by a rather substantial literature on lexical neighborhoods, where the "lexical neighborhood" of an existing or nonsense forms is defined by the set of words which differ in a single phoneme (according to the definition of Luce *et al.* 1990). Studies by Luce and colleagues demonstrate that the lexical neighborhood density of a word has a strong effect on word perception, which may be attributed to the number of active competitors for a word at each point in the speech signal. Studies relating lexical neighborhoods to acceptability include Ohala and Ohala (1986), who asked subjects to rate forms which violated an equal number of morpheme structure conditions, but which differed in their distance from actual words. The difference in ratings showed that the acceptability of a word was correlated with its distance from actual words, as proposed by Greenberg and Jenkins (1964), not with the number of MSC violations.

A smaller literature considers structural factors through intensive study of particular configurations. In a study of medial triconsonantal clusters, such as /lfr/ in "palfrey", Pierrehumbert (1994) showed that the independent probabilities of the coda and the following onset was the single biggest factor in predicting which complex clusters exist. Almost all of the 40 existing different triconsonantal clusters are among the 200 most probable if the complete cross-product of (frequency-tagged) onsets and codas is computed. Since the complete cross-product yields more than 8000 different candidate medial clusters, this is a very powerful factor. Results of an experiment described in that paper showed that subjects have an implicit awareness of the statistical underrepresentation of consonant sequences and reveal this awareness in judgments of well-formedness.

These two groups of papers leave many unanswered questions. Pierrehumbert (1994) provides no suggestions about how effects over the whole word may be combined. If Pierrehumbert's claim is extrapolated without elaboration, it entails that longer words should almost always be worse than shorter ones. Longer words, having more parts, would have more factors in their computed likelihoods, with each factor less than one (since the probability of any given choice is always less than one). Hence the longer the word, the more probable that its likelihood would be at a very low value. This difficulty is a classic problem for stochastic parsing, and it leads to suggestions about normalizing the scores. But a scoring system which completely normalized for length (e.g. by considering mean log probabilities) would provide no way of capturing the effect that Pierrehumbert reports, since the mean log probabilities of the nonexistent complex clusters would be no worse than the log probabilities of their component parts.

The lexical neighborhood literature also avoids the question of integration over the word, by virtue of threshholding on a single distance (obviously, a crude expedient adopted during a first pass at the problem). The question of how structure figures in the perceived relatedness of words has also not been taken up in the lexical neighborhood literature. The phoneme-wise calculation may be reasonably well-behaved if computed over monosyllables, but it is too crude a measure if the situation is considered in its full generality. For example, a single phoneme substitution which had a drastic effect on the syllable structure must surely yield a less cognitively related form than one which does not.

In order to advance our understanding of these issues, we have developed a probabilistic parser which handles the interactions amongst the following factors: 1) the phonemic content of the onset and of the rhyme; 2) the location with respect to the word edge; 3) the stress pattern within the word. These factors cover a substantial fragment of English phonotactics. We then parse a set of neologisms and compare various methods of scoring the goodness of the parse as a predictor of acceptability: 1) The overall acceptability of a form is the likelihood of the best parse. However, because long words contain more constituents than short words, their likelihood is lower, as more multiplications are involved. In order to offset this multiplicative effect, we also considered the following score: 2) The overall acceptability of a form is the log likelihood of the best parse. 3) The overall acceptability of the form is dominated by the worst component (the single lowest probability onset or rhyme). This alternative is loosely inspired by the phonological literature, from classical generative phonology to Optimality Theory, in which the badness of a form depends on its most egregious phonotactic constraint violation. 4) We also examined the idea that the overall acceptability of a form is dominated by the best constituent, in recognition of the experimental result that nonsense words such as "mrupation" are often not regarded by subjects as being particularly bad, since despite containing a very un-English onset, the remainder of the word, including its morphological and prosodic structures, are well-formed.

We find that of these four proposals for scoring phonotactic well-formedness, 1), 2) and 3) yield statistically significant correlations with experimentally obtained judgements.

## 2 Grammar and Parsing

For the present paper, we consider a grammar of English words which is extremely simple but which still offers enough complexity to cover a large fraction of the English vocabulary and to raise serious issues for a stochastic implementation. We consider all monosyllables and disyllables in Mitton (1992). Since these may differ in the stress of each syllable, this yields the following CF rules for expanding the word node W into strong and weak syllables Ss and Sw:

1) W → Ss       (monosyllabic words)
2) W → Sw Ss   (iambic words, such as "about")
3) W → Ss Sw   (trochaic words, such as "party")
4) W → Ss Ss   (words with two stresses, such as "Rangoon")

The disyllabic words in the dictionary also include quite a few compounds, which behave phonotactically like two monosyllabic words. In order to provide for such cases, the actual root node in the system is U ("utterance"), supporting expansions:

5) U → W
6) U → W W

Syllables have internal structure which is important for their phonotactics. According to classical treatments, such as Fudge (1969), each syllable has an onset and a rhyme, yielding the following rule schema:

7) S → O R

Some more recent theories of the syllable do not have onsets and rhymes as such, but distinguish the region of the syllable up to the head vowel from the region consisting of the head vowel and any following tautosyllabic consonants. The internal decomposition of the onset and the rhyme are highly controversial, with some theories positing highly articulated tree structures and others no structure at all. We sidestep this issue by taking onsets and rhymes to be unanalyzed strings. We adopted this approach because a prosodic grammar with two node levels is already sufficiently complex for our purposes, which is to compare the effects of local and diffuse phonotactic deviance.

One might think that rules 1) – 7), augmented by a large set of rules for spelling out the terminals, would provide a sufficient grammar to describe English monosyllabic and disyllabic words. But they do not. Difficulties arise because the inventories of onsets and rhymes are not the same at all positions in the word. Attempts to accommodate this fact provide a mainstay of the literature on syllabification. The main qualitative observations are the following: 1) Extra consonants are found at the end of the word which are non-existent or rare at the end of word internal syllables. The coronal affixes (/s/, /t/, and /θ/) provide the best known example of extra consonants. However, the pattern is much more pervasive, with many cases involving neither independent morphemes nor coronal consonants. Rhymes such as /ɛmp/ (as in "hemp") and /ælk/ as in "talc" are also more prevalent at the end of the word than in the middle. 2) Light syllables with a lax full vowel are permitted only nonfinally. 3) Word-initial syllables need not have an onset, whereas word-medial syllables usually have an onset (of at least one consonant): hiatus is uncommon.

Extraneous consonants at the words edges can be generated by supplementing a grammar of type 1) – 7) with rules such as 8).

8)      W → Ss C

As noted in McCarthy and Prince (1993), such a treatment fails to capture the fact that word edges provide a location for defective syllables in addition to overlarge ones. When we turn to probabilistic models, the limitations of the approach in 8) become even more apparent. The probability distributions for *all* onsets and rhymes depend on the position in the word. For example, /t/ is possible in coda position both finally (as in "pat") and medially (as in "jit.ney"). A classical grammar would stop at that. But a probabilistic grammar must undertake to the model the fact that /t/ is much more common as a word-final coda than as a word-medial one, and that acceptability judgments by native speakers reflect this fact (Pierrehumbert, 1994). Therefore, we handle deviance at the word edges in a different manner. Stochastic grammars provide us with the possibility of describing such effects by expanding (or rather, failing to collapse) the rules for subordinate nodes in the tree. Instead of attempting to assign a probability to rule 7), which applies regardless of the position of the syllable in the tree, we label the syllable nodes according to their position in the word, and propagate this labelling through all lower expansions. The total inventory of syllable types is then:

9)  Ssi    strong initial syllables which are not also final
    Ssf    strong final syllables which are not also initial
    Ssif   strong syllables which are both initial and final

and similarly for weak syllables, Swi, Swf and Swif. For a lexicon which included longer words, it would of course also be necessary to provide for medial syllables.

Propagating this type of indexing, we can then provide for the fact that the rhyme /ɛmp/ is more common word finally than elsewhere as follows:

10)    Ssf → Osf Rsf
       Rsf → "ɛmp",    p = X.

11)    Ssi → Osi Rsi
       Rsi → "ɛmp",    p = Y, where Y < X.

This is, obviously, a brute force solution to the problem. It has the penalty that it treats as unrelated cases which are, in fact, related. In order to allow monosyllabic words to display both word-initial anomalies for the onset, and word-final anomalies for the rhyme, it is necessary to posit the categories Ssif and Swif. But then the expansion of the Ssif rhyme becomes formally unrelated to that of the Ssf rhyme, and that of the Ssif onset is unrelated to that of the Ssi onset. The practical penalty is that proliferation of logically different types under this approach reduces the count of words which can be used in training the probabilities for any individual case. For the rarer cases, the result can be that the sample sizes are reduced to a point at which statistically reliable estimates of the probabilities are no longer available from a full-size dictionary.

This is a scientific problem in addition to an engineering problem. In developing robust and productive phonotactics, speakers must have a better ability than standard stochastic CFGs provide to treat different contexts as analogous so that data over these contexts can be collapsed together. In developing the present parser, we have made a further assumption which allows us to circumvent this problem. In general, the phonological effects of edges are concentrated right at the edge in question. This means that the effect of the left word edge is concentrated on the onset, while the effect of the right word edge is concentrated on the rhyme. The tabulation of probabilities can then be organized according to the vertical, root-to-frontier paths through the tree with only a highly restricted reference to the horizontal context. Specifically, we claim that the root-to-frontier paths are tagged only for whether the frontier is at the left and/or the right edge of the word. Some example paths, those of the word "candle", are:

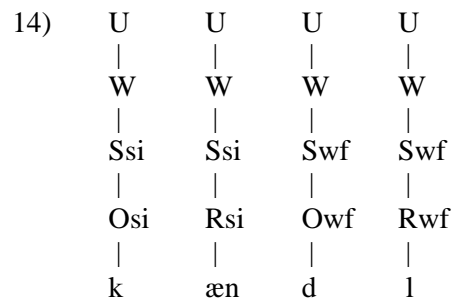

14)    U      U      U      U
       |      |      |      |
       W      W      W      W
       |      |      |      |
       Ssi    Ssi    Swf    Swf
       |      |      |      |
       Osi    Rsi    Owf    Rwf
       |      |      |      |
       k      æn     d      l

which we write for convenience U : W : Ssi : Osi : k, U : W : Ssi : Rsi : æn, etc.

Although the resulting representations are remiscent of those used in data-oriented parsing (see Bod, 1995), there is a very important difference. The paths we use partition the data;

each terminal string is an instance of only one path type, with the result that the probabilities add up to one over all paths. The result is that paths are properly treated as statistically independent, modulo any empirical dependencies which we have failed to model. DOP posits multiple descriptions which can subsume each other, so that any given syntactic fragment can contribute to many different descriptions. As a result, the descriptions are not independent by the very nature of the way they are set up.

To use the paths in parsing new examples, we zip consistent paths together from their roots downwards, unifying neighbouring categories as far down the paths as possible, an operation we call sequential path unification. The probability of the combined path is taken to be the product of the probabilities of the two parts. That is, since the original path set partitioned the data, a finite state model is a justifiable method of combining paths. Onsets and rhymes which are unattested in the original dictionary are assigned a nominal low probability by Good–Turing estimation (Good, 1953) which Bod (1995) argues to be better behaved than alternative methods for dealing with missing probability estimates for infrequent items.

The sequencing constraints described by the original grammar (for example, the requirement that an onset be followed by a rhyme and not by another onset) are enforced by tagging some nodes for the type of element which must succeed it, in a fashion reminiscent of categorial grammar. That is, onsets must be followed by rhymes with the same i/f and s/w subscripts, and initial syllables must be followed by final syllables, with an initial weak syllable followed by a strong syllable or an initial strong syllable followed by a weak one.

15) a) A successful instance of path unification

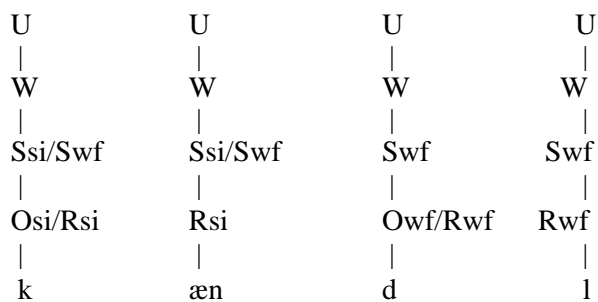

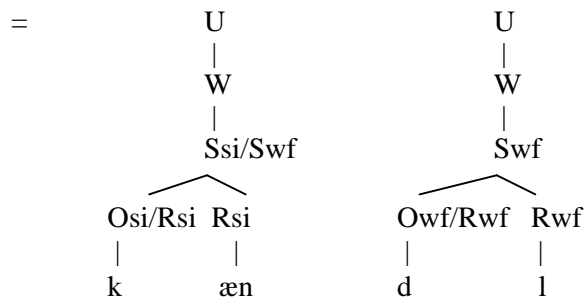

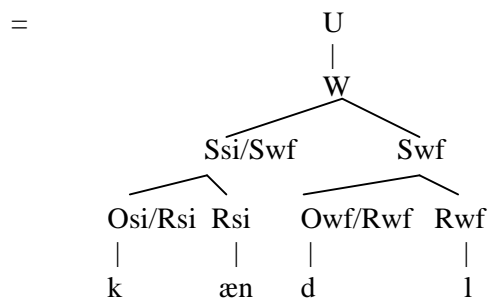

b) An unsuccessful attempt at path unification

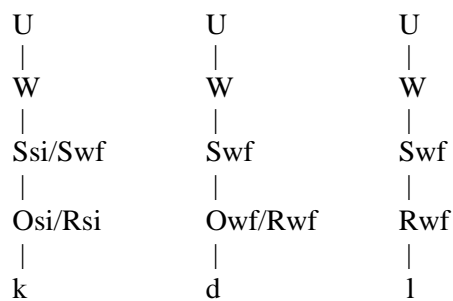

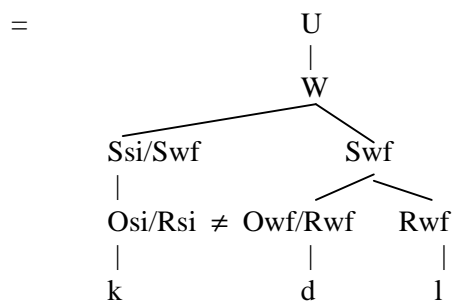

In 15b), the parse fails as the initial Osi is not followed by an Rsi, as it requires.

## 3 How the training was carried out

To establish the path probabilities for English monosyllabic and disyllabic words, the paths were tabulated over the 48,580 parsed instances of such words in Mitton (1992). With each word containing two to four paths, there was a total of 98,697 paths in the training set.

Parsing such a large set of words requires one to take a stand on some issues which are disputed in the literature. Here are the most important of these decisions. 1) We included every single form in the dictionary, including proper nouns, no matter how foreign or anomalous it might appear to be, because we have the working hypothesis that low probabilities can explain the poor productivity of anomalous patterns. 2) Following current phonological theory (see e.g. Ito 1988), we syllabified all word-medial VCV sequences as V.CV. As a related point, we took medial clusters beginning with /s/ to be syllable onsets when they were possible as word onsets. If the sC sequence is not an attested word onset, it was split medially (e.g. 'bus.boy").

There are a number of situations in which the dictionary does not mark phonological information which we know to be important. We have done our best to work around this fact, but in some cases our estimates are inevitably contaminated. Specifically: although compounds which are hyphenated in the dictionary can be (correctly) parsed as two phonological words, many compounds have no indication of their status and are parsed as if they were single words. Similarly, words # affixes such as -ly and -ness have been parsed as if they had no internal structure. This contaminates the counts for nonfinal rhymes with a certain number of final rhymes, and it contaminates the counts for noninitial onsets with a certain number of word-initial onsets. Second, stress is not marked in monosyllabic words. We have therefore taken all monosyllabic words to have a main word stress. As a result, a few reduced pronunciations for function words are included, with the result that there is a small, rather than a zero, probability for stressed syllable rhymes with a schwa. Third, secondary stresses are not reliably marked, particularly when adjacent to a primary stress (as in the word "Rangoon"). This means that a certain number of stressed rhymes have been tabulated as if they were unstressed. These problems for the most part can be viewed as sources of noise. We believe that the main trends of our tabulations are correct. To illustrate the fact that positional probabilities differ, table 1 compares the 10 most frequent onsets and rimes in each position.

Table 1.

| Osf | Osif | Osi | Owf | |
|---|---|---|---|---|
| s 234 | Ø 836 | Ø 1180 | l 979 | |
| t 206 | r 616 | k 848 | b 934 | |
| l 193 | b 614 | s 813 | t 884 | |
| r 164 | l 490 | p 767 | s 748 | |
| p 157 | k 489 | m 765 | Ø 746 | |
| m 152 | p 459 | b 725 | d 708 | |
| v 152 | t 453 | h 688 | n 698 | |
| f 139 | s 445 | t 627 | r 656 | |
| d 123 | h 444 | r 584 | m 621 | |
| k 123 | m 430 | l 567 | k 601 | |

| Rsf | Rsif | Rsi | Rwf | Rwi |
|---|---|---|---|---|
| eɪn 45 | i 365 | æ 950 | ɪ 740 | ɪ 815 |
| eɪt 41 | eɪ 147 | ɪ 819 | ɪz 703 | ə 742 |
| eɪts 37 | əʊ 114 | ɛ 694 | ɚ 661 | ɪn 203 |
| ɛt 37 | aɪ 107 | ɔ 654 | əz 644 | ən 120 |
| ɛs 34 | ʌn 95 | i 584 | l 514 | ʌn 87 |
| iz 34 | u 91 | eɪ 558 | lz 420 | ɒ 69 |
| ɛkt 33 | ʌp 89 | aɪ 537 | əs 417 | əʊ 60 |
| ɛkts 33 | eɪz 89 | ʌ 503 | ən 398 | ɪk 59 |
| ɛnt 33 | aʊt 88 | əʊ 472 | ə 226 | ɪm 59 |
| eɪ 32 | ɪn 76 | ɑ 429 | əʊ 213 | əb 49 |

## 4 Neologisms

The data set we used to evaluate the parser was obtained in a prior study (Coleman 1996). The goal of this study was to evaluate the psychological reality of phonotactic constraints. The materials were designed to permit minimal comparisons between a nonsense word which was in principle possible and one which was expected to be impossible by virtue of containing an onset or a rhyme which does not occur at all in the Mitton (1992) dictionary. Thus, the materials were made up of paired words such as /ˈmlɪsləs/ (impossible by virtue of the cluster /ml/) and /ˈglɪsləs/ (otherwise identical, but containing the attested cluster /gl/ instead of /ml/).

The materials were randomized, with a post-hoc test to ensure that related items in a pair were separated in the presentation. The words were recorded by John Coleman and presented aurally, twice over, to 6 naive subjects, who judged whether each word could or could not be a possible English word by pressing one of two response buttons. The total number of responses against the well-formedness of each word was taken as a score of subjective degree of well-formedness.

The distributions of scores of forms containing non-occuring clusters and those containing occuring clusters were significantly distinct. Forms which were designed to be "bad" were judged significantly worse than forms which were designed to be "good". This was the case for the pooled data, and for each matched pair, the "bad" variant received a lower score than "good" variant for 61/75 pairs. However the data contained a number of surprises, some of which, indeed, motivated the present study. The scores of the "bad" forms were much more variable than anticipated. "Bad" forms in some pairs (e.g. /mɹuˈpeɪʃn/) were scored better than "good" forms in other pairs (e.g. /ˈsplɛtɪˌsɑk/). Apparently, a single subpart of zero (observed) probability is not enough to render a form impossible. Conversely, forms which violate no constraints, but which are composed of low frequency constituents and have few lexical neighbors, are assigned low acceptability scores e.g. /ˈfɪŋkslʌp/ and /ʃəˈlɛnð/, which scored 12, i.e. completely unacceptable.

These findings are contrary to the predictions both of a classical phonological treatment (according to which linguistic competence is categorical, and forms which cannot be parsed are impossible) as well as to the predictions of Optimality Theory (in which a single severe deviation should determine the evaluation of the form). Apparently, the well-formed subparts of an otherwise ill-formed word may alleviate the ill-formed parts, especially if their frequency is high, as in the "ation" part of "mrupation" (/mɹuˈpeɪʃn/).

We used the stochastic grammar to parse the 116 mono- and di-syllabic neologisms from the earlier study, and compared various methods of scoring the goodness of the parse as a predictor of the experimentally obtained measure of acceptability. Specifically, we compared the four alternatives discussed in the introduction. Of the four proposals for scoring phonotactic well-formedness, three yield statistically significant correlations with experimentally obtained judgements. (Significance was assessed via a t-test on $r^2$, two-tailed, df = 114.)

| Scoring method | Significance of correlation |
|---|---|
| 1) p(word) | $p < .01$ |
| 2) ln(p(word)) | $p < .001$ |
| 3) p(worst part) | $p < .01$ |
| 4) p(best part) | n.s. |

Scoring method 2) is a better model of acceptability than 1) because it linearizes the exponential shape of p(word) arising from the multiplication of successive parts. Figure 1 is a scatterplot of the best correlation, ln(p(word)) against the number of votes against well-formedness. It is apparent that less probable words are less acceptable.

Figure 1.

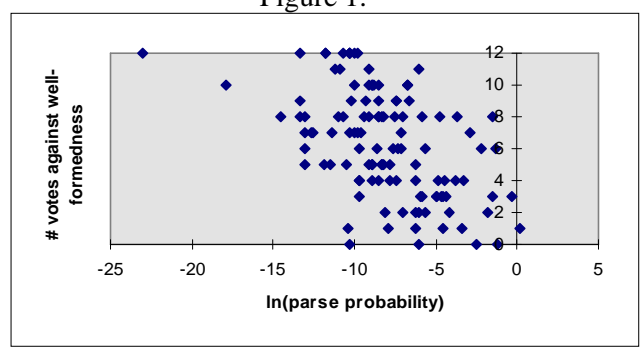

## 5 Discussion and Conclusions

We have compared several methods of using frequency information to predict the acceptability of neologisms. Both the probability of the word and the probability of the worst part are significant correlates of acceptability. This finding is significant, because the single worst violation dominates the determination of well-formedness in almost all varsions of generative phonology. In Chomsky and Halle (1968), morpheme structure conditions act as a filter on underlying representations. The same concept of grammatically is proposed in approaches founded on Boolean logic, such as Declarative Phonology. According to Optimality Theory, "impossible words" are those in which a constraint is so strong that a null parse is prefered to a parse in which the constraint is violated. This means that impossible words are those which are egregious according to a single constraint.

However, the probability of the worst part is not the best score of acceptability: the log probability of the whole word is a better measure, a result at odds with standard generative phonology and OT alike. In classical generative phonology, a UR which violates any single morpheme structure condition is ruled out absolutely. In more recent versions of generative phonology which build prosodic structure through some version of parsing or template mapping, the entire parse fails if it fails at any single point. The same idea shows up in a new guise in Optimality Theory. According to Optimality Theory, constraint violations do not interact cumulatively. A rank-ordering of constraints has the consequence that weak constraints can be violated to meet stronger ones, but there is no mechanism by which adherence to many weak constraints ameliorates the effect of a single violation of a stronger constraint. Our results indicate that these models achieve some success, but miss an important fact: the well-formedness of lexically attested parts ameliorates the unacceptability of the unattested or low-frequency parts. When statistically valid data on acceptability is gathered (as against the isolated intuitions of individual researchers/authors), it is found that deviations are partially redeemed by good parts, and that forms which are locally well-formed, in the sense that each piece is reasonably well-attested, can nonetheless be viewed as improbable overall. This finding supports the view that phonotactic constraints are probabilistic descriptions of the lexicon, and that probabilisitic generative grammars are a more psychologically realistic model of phonological competence than standard generative phonology and Optimality Theory.

**Acknowledgments**

This work was supported by NSF grant number BNS-9022484 to Northwestern University, by a fellowship from the John Simon Guggenheim Memorial Foundation to Janet Pierrehumbert, and by the University of Oxford.